\documentclass[11pt]{article}

\usepackage[utf8]{inputenc}
\usepackage[T1]{fontenc}
\usepackage{amsmath,amssymb,amsfonts}
\usepackage{graphicx}
\usepackage{hyperref}
\usepackage{booktabs}
\usepackage{color}
\usepackage{xcolor}
\usepackage{authblk}
\usepackage{caption}
\usepackage{subcaption}

\usepackage{booktabs}  
\usepackage{array}     

\usepackage[symbol]{footmisc}

\usepackage{nameref}

\usepackage{lscape}
\usepackage{geometry}
\usepackage[numbers]{natbib}
\title{KNOWLEDGE GRAFTING: A Mechanism for Optimizing AI Model Deployment in Resource-Constrained Environments}

\author[1]{Osama Almurshed\thanks{Authors contributed equally in this work.}}
\author[2]{Ashish Kaushal\textsuperscript{*}}
\author[3]{Asmail Muftah}
\author[2]{Nitin Auluck}
\author[4]{Omer Rana}

\affil[1]{Prince Sattam Bin Abdulaziz University, Kingdom of Saudi Arabia}
\affil[2]{Indian Institute of Technology Ropar, India}
\affil[3]{Azzaytuna University, Libya}
\affil[4]{Cardiff University, United Kingdom}

\date{}

\begin{document}

\maketitle

\begin{abstract}

The increasing adoption of Artificial Intelligence (AI) has led to larger, more complex models with numerous parameters that require substantial computing power -- resources often unavailable in many real-world application scenarios. Our paper addresses this challenge by introducing \textit{knowledge grafting}, a novel mechanism that optimizes AI models for resource-constrained environments by transferring selected features (the scion) from a large donor model to a smaller rootstock model. The approach achieves an 88.54\% reduction in model size (from 64.39 MB to 7.38 MB), while improving generalization capability of the model. Our new rootstock model achieves 89.97\% validation accuracy (vs. donor's 87.47\%), maintains lower validation loss (0.2976 vs. 0.5068), and performs exceptionally well on unseen test data with 90.45\% accuracy. It addresses the typical size vs performance trade-off, and enables deployment of AI frameworks on resource-constrained devices with enhanced performance. We have tested our approach on an agricultural weed detection scenario, however, it can be extended across various edge computing scenarios, potentially accelerating AI adoption in areas with limited hardware/software support -- by mirroring in a similar manner the horticultural grafting enables productive cultivation in challenging agri-based environments.
\end{abstract}


\section{Introduction}
\label{sec:introduction}

The exponential growth in complexity and size of artificial intelligence (AI) models has transformed their capabilities but simultaneously created significant deployment challenges. Modern neural networks with tens or hundreds of millions of parameters demand substantial computational resources -- a requirement that is in contrast with the limited capabilities of many real-world deployment frameworks \cite{canziani2016analysis}. This growing disconnect creates a technological divide, where AI remains inaccessible to a wide range of applications operating in resource-constrained environments.

Recent years have witnessed remarkable advancements in AI capabilities, along-with tremendous growth in model sizes. State-of-the-art neural networks like VGG16 and transformer-based architectures require millions or billions of parameters \cite{simonyan2014very, brown2020language}, imposing severe computational and memory demands that exceed the capabilities of most edge devices, embedded systems, and mobile platforms.

The growing gap between model complexity and available computational resources creates several interrelated challenges: prohibitive costs limiting adoption to well-resourced institutions \cite{strubell2019energy}; concerning carbon footprints equivalent to the lifetime emissions of multiple cars \cite{lacoste2019quantifying}; and restricted access to AI capabilities in resource-constrained environments like IoT applications where the potential benefits are substantial \cite{almurshed2022adaptive}. Critical bottlenecks preventing widespread deployment include: excessive memory requirements beyond available RAM on edge devices \cite{howard2017mobilenets}; computational complexity leading to unacceptable latency on devices without specialized AI accelerators \cite{liang2021pruning}; and energy constraints that limit feasibility for battery-powered devices \cite{yang2017designing}.

Various techniques have been developed to address these challenges. Quantization reduces numerical precision to decrease memory and computational requirements \cite{jacob2018quantization}, but often introduces accuracy degradation, particularly for aggressive bit-width reduction \cite{han2015deep}. Pruning selectively removes less important connections or neurons \cite{guo2016dynamic, han2015learning}, but typically requires careful retraining to maintain accuracy. Knowledge distillation trains a smaller ``student'' model to mimic a larger ``teacher'' model \cite{hinton2015distilling}, but struggles with high compression ratios \cite{polino2018model}. Transfer learning applies pretrained layers to new tasks \cite{kornblith2019better, yosinski2014transferable}, but often preserves the computational complexity of original models.

This paper introduces \textit{knowledge grafting} -- an approach inspired by horticultural practices where gardeners have solved similar optimization problems for centuries. In horticulture, grafting combines desirable traits from different plants into more productive hybrids that thrive in challenging environments. Similarly, our knowledge grafting technique selectively transfers critical features (``scions'') from comparatively large models to smaller, more efficient ``rootstock'' models. Unlike existing techniques that primarily modify existing architectures through reduction (pruning) or compression (quantization); knowledge grafting fundamentally redesigns model architecture by selectively transferring intermediate features rather than entire layers, creating hybrid architectures that combine the best aspects of different models, prioritizing what to keep rather than what to remove, and also eliminating complex retraining procedures typically required by knowledge distillation.

We also formalized the grafting process through a mathematical framework which shows the systematic feature selection and transfer with dual optimization objectives: size-constrained performance maximization or performance-constrained size minimization. The results demonstrate the effectiveness of our knowledge grafting technique -- achieving an 88.54\% reduction in model size, while maintaining 89.97\% validation accuracy. The technique is designed with an aim to allow AI frameworks to be deployed on resource-constrained devices, without compromising performance. This will open new possibilities for AI applications in environments which were previously considered too limited to utilize machine learning frameworks. The main highlights of this paper are as follows:-
\begin{itemize}
    \item Introduce a new \textit{Knowledge Grafting} technique that transfers selected (important) features from large donor models to smaller rootstock architectures.
    
    \item Design a mathematical framework for systematic feature selection with dual optimization objectives.
    
    \item Achieve an 88.54\% model size reduction (64.39MB to 7.38MB), while also improving generalization performance of the model.
    
    \item Achieve better rootstock model performance (89.97\% validation accuracy), compared to donor model (87.47\%) with consistent test accuracy of 90.45\%.
    
    \item Obtain 8.7x size reduction, outperforming quantization (2x - 4x), pruning (3x - 5x), and knowledge distillation (5x - 6x).
\end{itemize}

The rest of the paper is structured as follows: First, we establish the theoretical foundations of Knowledge Grafting (Section \ref{sec:feature_grafting}), addressing the challenge of developing efficient AI models. We then define a use case scenario of agricultural weed detection that we use throughout the paper (Section \ref{sec:agricultural_use_case}). A mathematical formulation (Section \ref{sec:mathematical_problem}) and experimental methodology (Section \ref{sec:experimental_design}) are provided, followed by a comprehensive results analysis (Section \ref{sec:experimental_results}). This analysis covers training dynamics, testing performance, and notably, significant model size reduction. To contextualize our approach, we compare grafting against state-of-the-art models (Section \ref{sec:comparative_analysis}), and review related optimization techniques (Section \ref{sec:related_work}). The paper concludes by proposing a road map and future directions for this method in AI (Section \ref{sec:future_research}) and establishing Knowledge Grafting as a promising paradigm for resource-efficient AI that maintains high performance (Section \ref{sec:conclusion}).

\begin{figure}[!h]
    \centering
    \includegraphics[width=\linewidth]{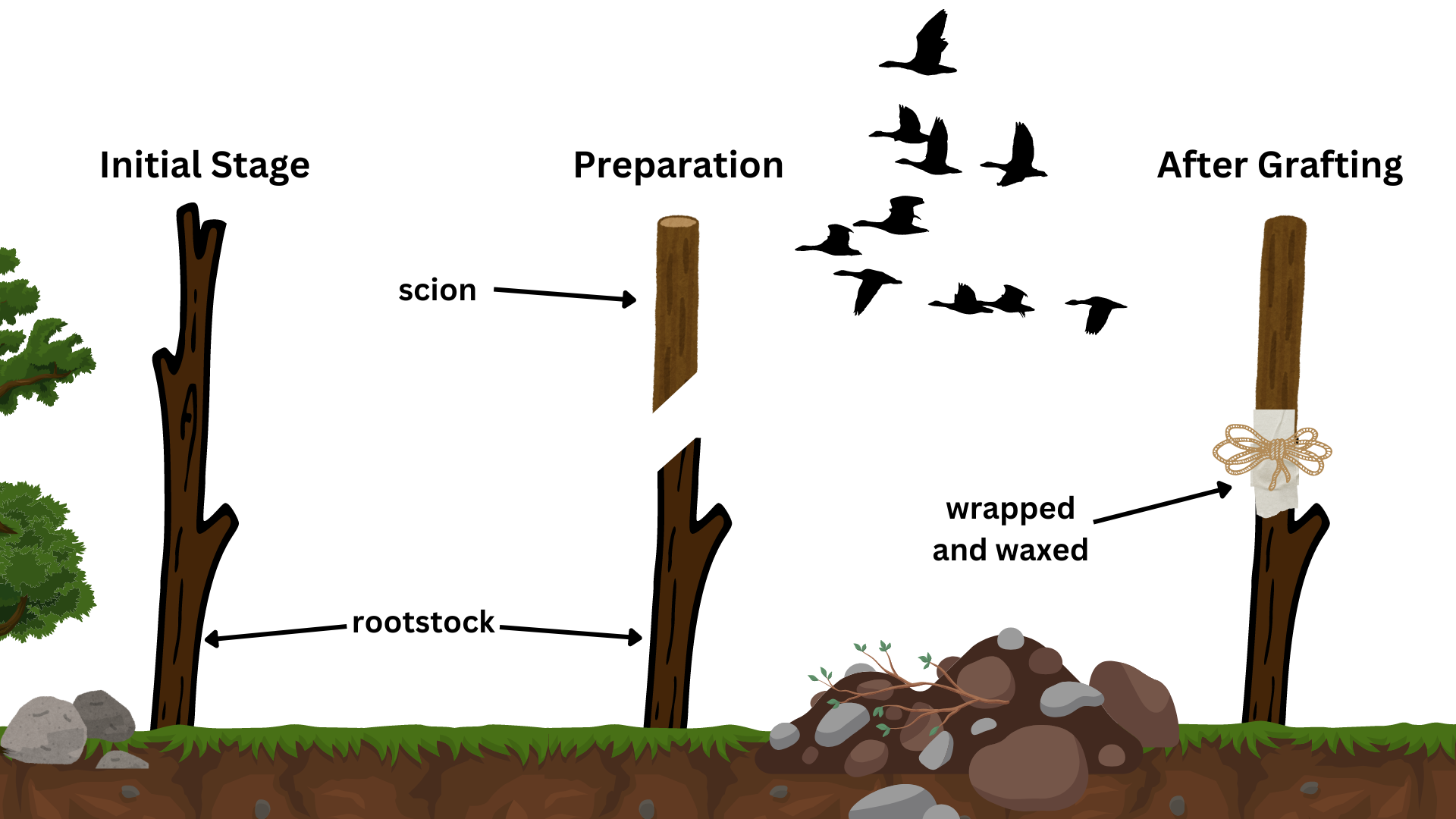}
    \caption{A depiction of grafting process in plants.}
    \label{fig:grafting_plants}
\end{figure}

\section{Knowledge Grafting for Artificial Intelligence}
\label{sec:feature_grafting}

We introduce Knowledge Grafting as an innovative approach in deep learning that draws inspiration from horticultural techniques practiced for centuries. As illustrated in Fig~\ref{fig:grafting_plants}, traditional plant grafting combines two plant parts: a robust rootstock providing a strong foundation, and a carefully selected scion bearing desired characteristics. This biological union creates a plant possessing the best attributes of both sources. In traditional agriculture, grafting allows growers to combine disease resistance and environmental adaptability with superior productive qualities, enabling cultivation of desirable varieties in conditions where they would otherwise struggle to survive.

In our Knowledge Grafting approach for neural networks, we apply the similar biological principle to artificial intelligence. We begin by cultivating a donor model using an architecture like VGG16 pretrained on a dataset. This donor model develops rich learned features, but proves too computationally intensive for deployment in resource-constrained environments. Rather than attempting to compress this entire model, we identify and select specific intermediate layers containing the most valuable learned features-- which we call ``scions.'' These layers have developed sophisticated feature extraction capabilities through extensive training that would be difficult to replicate in a smaller model trained from scratch. For the rootstock, we develop a lightweight model architecture specifically designed for efficient computation. This model serves as the foundation that will receive the grafted features, prioritizing computational efficiency over complex feature extraction. The key innovation occurs at the grafting union, where we extract features from selected layers of the donor model using Global Average Pooling. These transformed features are then grafted onto our rootstock model by concatenating them and connecting them through newly added dense layers that integrate these rich features into the efficient architecture.

This methodology differs significantly from existing model optimization techniques described in previous sections. Unlike pruning, which simply removes parts of an existing model, our approach selectively transfers the most valuable components. Furthermore, while quantization reduces numerical precision, our approach maintains full precision of critical features. Additionally, in contrast to knowledge distillation, Knowledge Grafting directly transfers well-developed features without lengthy retraining, addressing the limitations of conventional transfer learning approaches mentioned earlier in this work.
\begin{figure}[h]
    \centering
    \includegraphics[width=\linewidth]{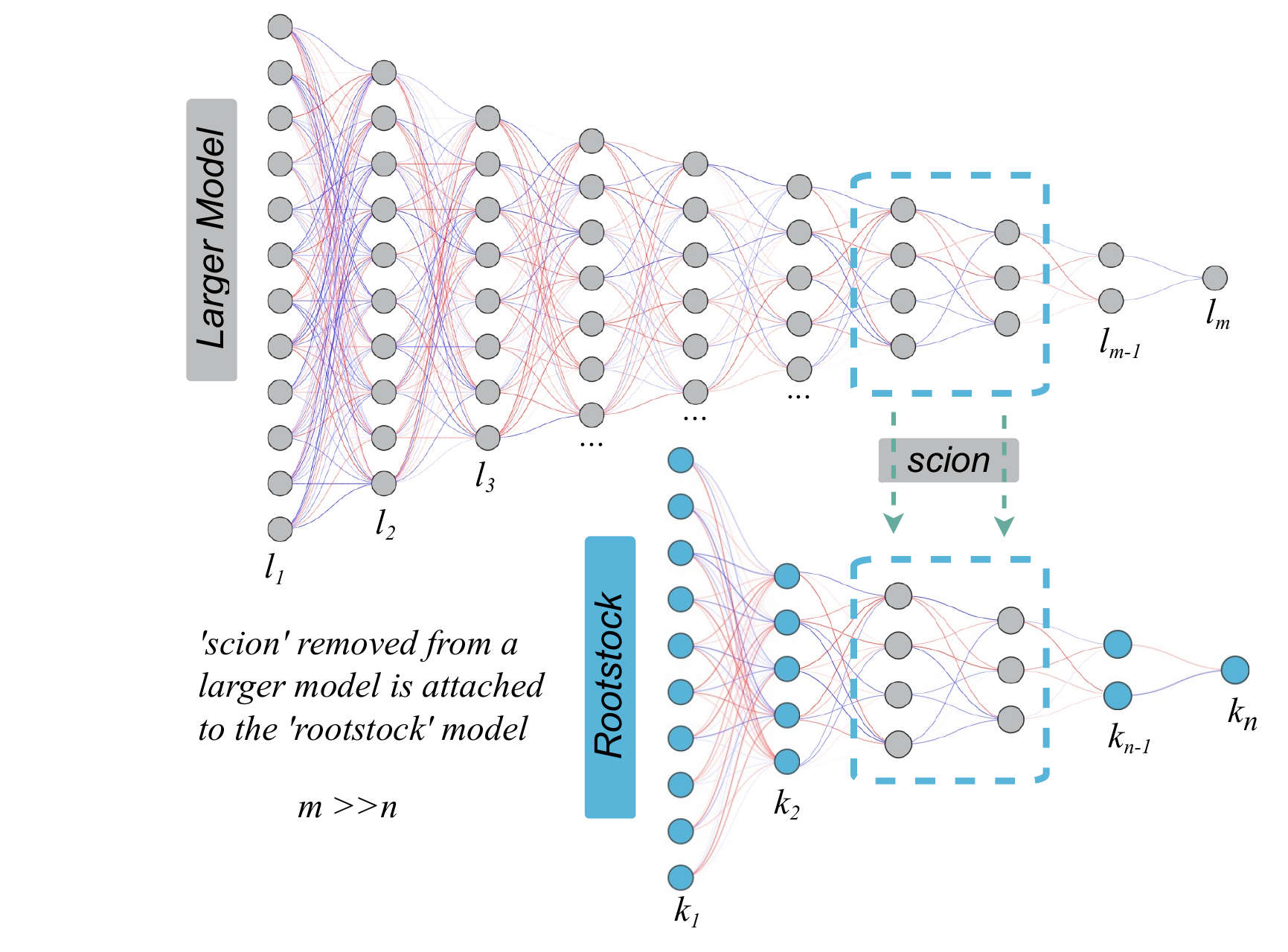}
    \caption{Grafting process in a neural network.}
    \label{fig:nnGraphting}
\end{figure}

Fig~\ref{fig:nnGraphting} illustrates how this biological inspiration translates into the neural network architecture. The upper model shows the donor model with its comprehensive set of layers, while highlighted components represent the selected features that will serve as our scion. The lower model displays the resulting grafted model, demonstrating how these valuable features are integrated into the more efficient rootstock architecture. Through Knowledge Grafting, we plan to enable AI models to benefit from large model learning, without inheriting their full resource requirements. This is particularly important for applications like the weed detection system described in later sections, where computational resources are severely constrained. Our approach represents a promising direction in deep learning, offering a pathway to propagate the advantages of large-scale models into practical, efficient implementations suitable for environments with limited computational resources.

\section{Agricultural Use Case: Weed Detection}
\label{sec:agricultural_use_case}

The real-world application of knowledge grafting in resource-constrained environments can be precision agriculture where efficient AI deployment can significantly enhance productivity, while facing substantial technical limitations. In modern precision agriculture, autonomous robots have emerged as powerful tools for field management tasks, especially weed detection and elimination. These systems typically uses camera-equipped robots that methodically traverse agricultural fields, capturing and analyzing images to differentiate between crops and unwanted weeds \cite{patros2022rural}. Upon successful identification of a weed, the robot promptly initiates a targeted removal process before continuing its systematic survey of the field.

Effective weed species identification presents technical challenges particularly when deploying sophisticated deep learning models on field-based devices with limited computational capacity. The DeepWeeds dataset \cite{DeepWeeds2019}, comprising 17,509 images of eight weed species collected across northern Australia, has proven valuable for training classification models; however, the computational requirements of state-of-the-art architectures pose significant deployment barriers. Models such as VGG16 and InceptionV3 can achieve impressive accuracy in weed classification, but require millions of parameters and billions of operations to process a single image, rendering them impractical for direct deployment on field robots with constrained resources~\cite{kaushal2023edge}.

Agricultural robots operate under several critical constraints that further complicate the deployment of AI models. These robots rarely perform image classification in isolation; they simultaneously monitor soil conditions, coordinate with nearby robots, collect various sensor data, and manage navigation systems \cite{almurshed2022adaptive}. This multifaceted operation makes it impractical to deploy computation-heavy ML models that would monopolize the limited computational resources available on-board. Additionally, rural agricultural environments frequently suffer from poor network infrastructure characterized by unreliable wireless connectivity and restricted bandwidth \cite{kaushal2024shield}. The hardware platforms utilized in agricultural settings typically lack specialized computational resources such as dedicated AI accelerators, leading to unacceptable latency and reduced operational effectiveness. Energy efficiency also represents a significant constraint, with larger models requiring substantially more power -- as observed in the ToSiM-IoT framework, where models can consume significant wattage per training epoch \cite{kaushal2024tosim}.

Given these constraints, effective AI deployment in agricultural robotics necessitates models with minimal size footprints. With restricted data transmission capabilities in rural areas, reducing the volume of data required for model updates is essential for operational continuity. Compact models require less power for transmission, and enable more efficient computation on general-purpose processors \cite{howard2017mobilenets}. Feature selection techniques are particularly valuable in this context where reducing retained features to 10\% decreased training time by 66.26\% for the VGG16 model \cite{kaushal2024tosim}. More efficient models also reduce power consumption and associated carbon emissions, aligning with broader sustainable development goals in modern agriculture.

These specific constraints and requirements in agricultural robotics create an ideal application scenario for our knowledge grafting approach. By selectively transferring only the most critical features from a comprehensive donor model to a lightweight rootstock model, we can create highly efficient AI systems that maintain detection accuracy, while dramatically reducing computational requirements. This technique enables effective deployment on resource-constrained agricultural robots, without compromising their ability to perform multiple concurrent tasks essential for productive field operations.

In the next section, we formalize the optimization challenge as a mathematical problem providing a systematic framework for selecting which features to transfer from donor to rootstock models to achieve maximum efficiency, while maintaining the high performance necessary for reliable weed detection in agricultural environments.

\section{Mathematical Problem Definition}
\label{sec:mathematical_problem}

The Knowledge Grafting process can be formally defined with a mathematical framework which captures both the selection mechanism and the optimization objectives. This formalization provides a description of how features are transferred from the comprehensive donor model to the more efficient rootstock model. We define the donor model (larger model in Fig~\ref{fig:nnGraphting}) as $D$, comprising of layers $\{l_1, l_2, ..., l_m\}$. When presented with an input $x$, each layer $l_i$ produces a feature map output denoted as $f_i(x)$. Similarly, the rootstock model contains layers $\{k_1, k_2, ..., k_n\}$, where typically $m \gg n$, showing the desired reduction in model complexity. The core of knowledge grafting involves selecting which layers from the donor model will serve as the ``scion'' to be grafted onto the rootstock. This selection can be represented as a binary vector:

\begin{equation}
s = [s_1, s_2, ..., s_m]
\end{equation}

Here, $s_i \in \{0,1\}$ indicates whether layer $i$ is selected ($s_i = 1$), or not ($s_i = 0$). The set of selected layers, constituting our scion can be represented as:
\begin{equation}
S = \{i \mid s_i = 1, i \in \{1,2,...,m\}\}
\end{equation}

For each selected layer in the scion, we apply a transformation function $T$ (Global Average Pooling, as discussed in Section~\ref{sec:experimental_design}), to convert the feature maps into a more compact representation:

\begin{equation}
g_i(x) = T(f_i(x)) \text{ for all } i \in S
\end{equation}

These transformed features are then combined using a composition function $C$ (concatenation), and follows the process of attaching the scion to the rootstock:

\begin{equation}
h(x) = C(\{g_i(x) \mid i \in S\})
\end{equation}

Finally, the combined features are mapped to the output space through a function $M$ (by implementing dense layers with activation functions like LeakyReLU, as outlined in Section~\ref{sec:experimental_design}):

\begin{equation}
y(x) = M(h(x))
\end{equation}

The resulting grafted model $G$ is defined by the selection vector $s$, along with the transformation, composition, and mapping functions, is given as $G(x)$:

\begin{equation}
G(x) = M(C(\{T(f_i(x)) \mid i \in S\}))
\end{equation}

This formulation creates a combinatorial selection problem with $2^m$ possible configurations. Given that the exhaustive search would be computationally infeasible for practical neural networks as $m$ can be large, we can approach this optimization problem in two distinct ways:

The first optimization objective can be represented as Size-Constrained Performance Maximization such that we aim to:

\begin{equation}
\text{Find } s^* = \arg\max_s P(G_s)
\end{equation}
\[
\text{subject to: } \text{Size}(G_s) \leq \text{Size}_{\text{max}}
\]

This objective function seeks to find the configuration that maximizes model performance, while ensuring the model remains below a predetermined size threshold. This approach is particularly relevant for edge devices with strict hardware limitations, such as the agricultural robots (described in Section~\ref{sec:agricultural_use_case}) where computational resources and memory are finite and constrained. The second optimization objective can be given as Performance-Constrained Size Minimization, such that:

\begin{equation}
\text{Find } s^* = \arg\min_s \text{Size}(G_s)
\end{equation}
\[
\text{subject to: } P(G_s) \geq P_{\text{min}}
\]

This objective function aims to minimize the model size, while maintaining a minimum acceptable performance level. This approach is especially valuable in bandwidth-constrained environments common in rural agricultural settings, where model updates must be transmitted efficiently over limited network connections.

Furthermore, $P_{\text{min}}$ defines the minimum acceptable performance threshold that ensures the model remains useful for its intended application. For weed detection in precision agriculture, this threshold must be high enough to reliably distinguish between crops and weeds to minimize both missed detections and false positives. The parameter $\text{Size}_{\text{max}}$ establishes the maximum allowable resource usage, determined by the hardware constraints of the target deployment platform. For agricultural robots operating in the field, this constraint is particularly stringent due to limited onboard computing capabilities and the need to balance multiple concurrent tasks including navigation, sensor data collection, and operational control.

\begin{table}[h]
\centering
\caption{Mathematical notation used in knowledge grafting formulation.}
\label{tab:notations}
\begin{tabular}{|c|p{12cm}|}
\hline
\textbf{Notation} & \textbf{Description} \\
\hline
$D$ & Donor model (larger pretrained model) with layers $\{l_1, l_2, ..., l_m\}$ \\
\hline
$f_i(x)$ & Feature map output from layer $i$ of donor model for input $x$ \\
\hline
$s$ & Binary selection vector $[s_1, s_2, ..., s_m]$ indicating which layers are selected \\
\hline
$S$ & Set of selected layer indices that form the scion \\
\hline
$T$ & Transformation function (e.g., Global Average Pooling) applied to selected features \\
\hline
$g_i(x)$ & Transformed feature from layer $i$ \\
\hline
$C$ & Composition function (typically concatenation) to combine transformed features \\
\hline
$M$ & Mapping function (implemented as dense layers) to produce final output \\
\hline
$G(x)$ & Resulting grafted model defined by selection $s$ \\
\hline
$P(G_s)$ & Performance metric (e.g., validation accuracy) of grafted model \\
\hline
$\text{Size}(G_s)$ & Resource usage metric (e.g., parameter count, model size) of grafted model \\
\hline
$P_{\text{min}}$ & Minimum acceptable performance threshold for the model \\
\hline
$\text{Size}_{\text{max}}$ & Maximum allowable resource usage for deployment \\
\hline
\end{tabular}
\end{table}

The selection of the appropriate objective function depends on the specific deployment scenario and priorities. In our agricultural use case, the performance-constrained size minimization approach proved most appropriate as it allowed us to significantly reduce model size, while maintaining the high accuracy essential for proper weed identification and treatment. The mathematical framework presented here not only formalizes the knowledge grafting concept, but also establishes a foundation for systematic optimization of the selection process, potentially through genetic algorithms or other heuristic approaches that could efficiently navigate the vast combinatorial space of possible configurations. Our implementation, as detailed in Section~\ref{sec:experimental_design}, demonstrates the practical application of this framework, achieving substantial model reduction, while preserving performance for agricultural applications.

\section{Experimental Design}
\label{sec:experimental_design}

To design our framework, we first prepare our donor model using a VGG16 architecture pretrained on ImageNet. We then prune the top layers and graft on custom dense layers with LeakyReLU activations and dropout for regularization, creating the donor model. This donor model is then trained on our specific dataset. For the rootstock model, we select specific layers from the donor model -- in this case, layers 8, 9, and 10 -- as our grafting points. These layers are chosen for their rich, intermediate feature representations to be the scion. We extract these features using Global Average Pooling, effectively trimming them down to their essential characteristics. These pooled features are then grafted onto our rootstock model by concatenating them and feeding them through new dense layers. This grafting process allows the rootstock to benefit from the donor model's learned features, while maintaining a smaller, more efficient structure. Both models are trained using the Adam optimizer with categorical cross-entropy loss, and their performance is carefully monitored and visualized to ensure successful feature transfer and overall model health. The Knowledge Grafting Process is divided into four main steps:

\begin{enumerate}
\item \textbf{Donor Model Cultivation:} This stage involves preparing the base model. A pretrained VGG16 model is loaded without its top layers. The early layers, up to the 17th from the end, are frozen. The model is then modified by adding a Flatten layer, two Dense layers with 256 units each using LeakyReLU activation, Dropout layers for regularization, and a final Dense layer with softmax activation. The model is compiled using Adam optimizer and categorical cross-entropy loss, then trained on the dataset.

\item \textbf{Data Preparation:} In this stage, image data is loaded and preprocessed. The pixel values are normalized, and labels are encoded using LabelEncoder and one-hot encoding. The data is split into training (60\%), validation (20\%), and testing (20\%) sets. A custom data generator is created for batch processing.

\item \textbf{Scion Creation:} Specific layers (8, 9, and 10) are selected from the rootstock model. The output from each of these layers is extracted and subjected to Global Average Pooling. These pooled outputs are then concatenated. A Dense layer with 256 units and ReLU activation is added followed by a Dropout layer and a final Dense layer with softmax activation. The scion model is compiled using Adam optimizer and categorical cross-entropy loss.

\item \textbf{Grafting \& Training:} 
The rootstock model training process uses a batch size of 16 and ran for 18 epochs. The grafted model combining donor features (scion) with the lightweight architecture (rootstock), used the same data generators as the donor model to ensure consistency. Throughout training, metrics including loss and accuracy for both training and validation sets were monitored each epoch to evaluate learning dynamics and ensure successful feature transfer.
\end{enumerate}

\section{Experimental Results and Analysis}
\label{sec:experimental_results}

\begin{figure*}[h]
    \centering
    \includegraphics[width=0.48\linewidth]{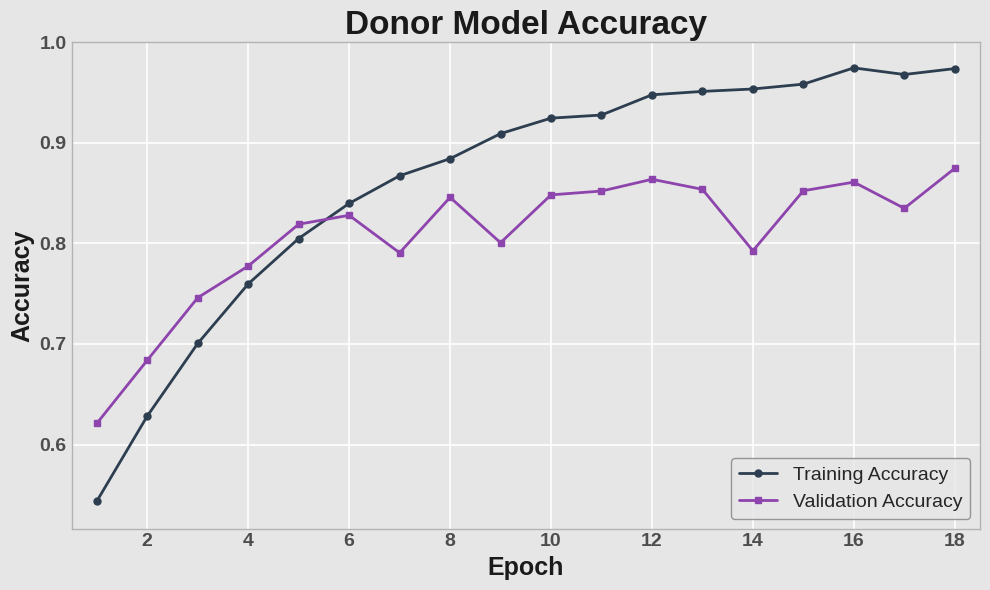} \hfill
    \includegraphics[width=0.48\linewidth]{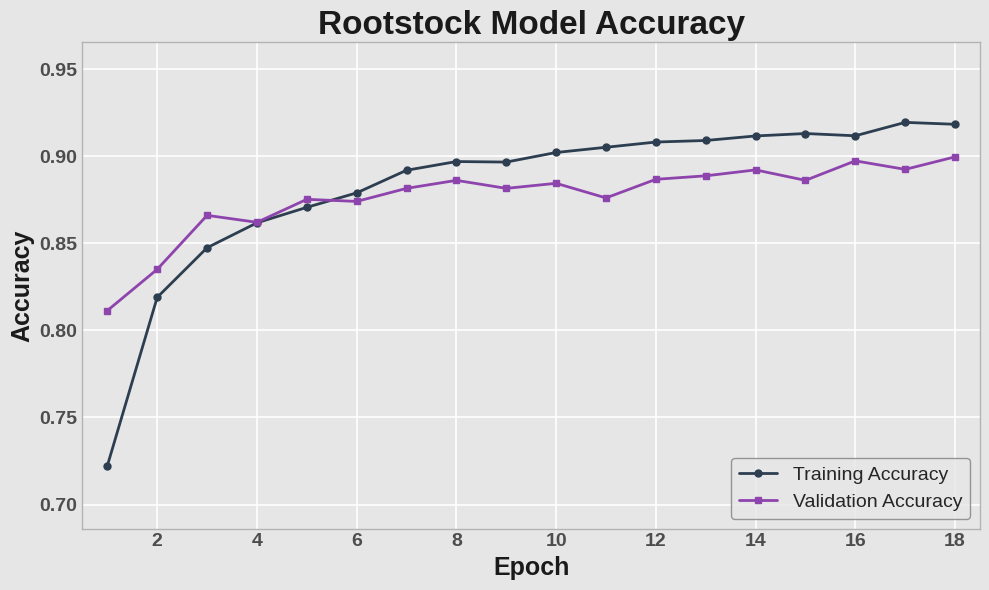} \\
    (a) Donor model accuracy \hfill (b) Rootstock model accuracy \\[0.2cm]
    \includegraphics[width=0.48\linewidth]{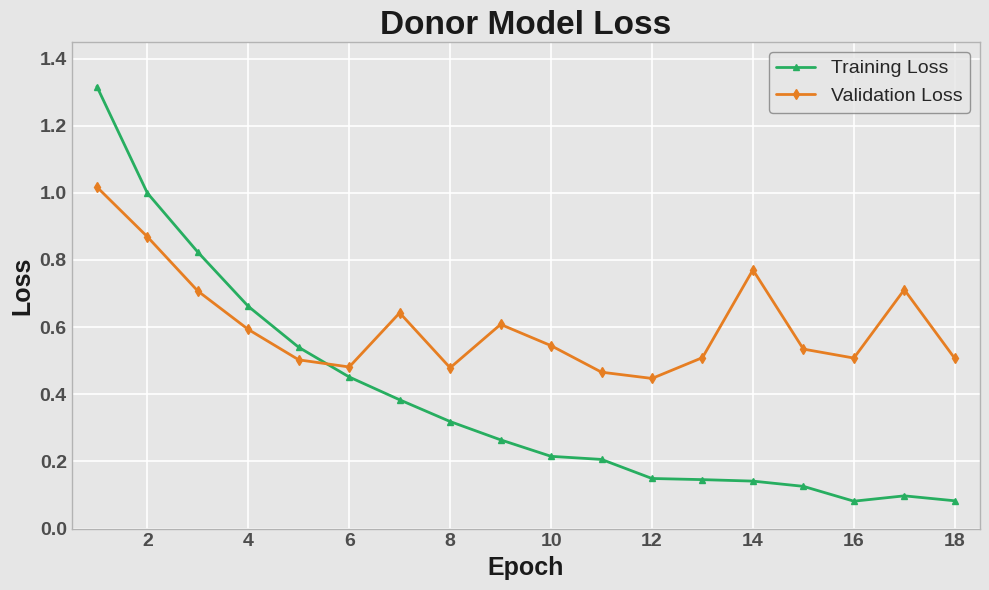} \hfill
    \includegraphics[width=0.48\linewidth]{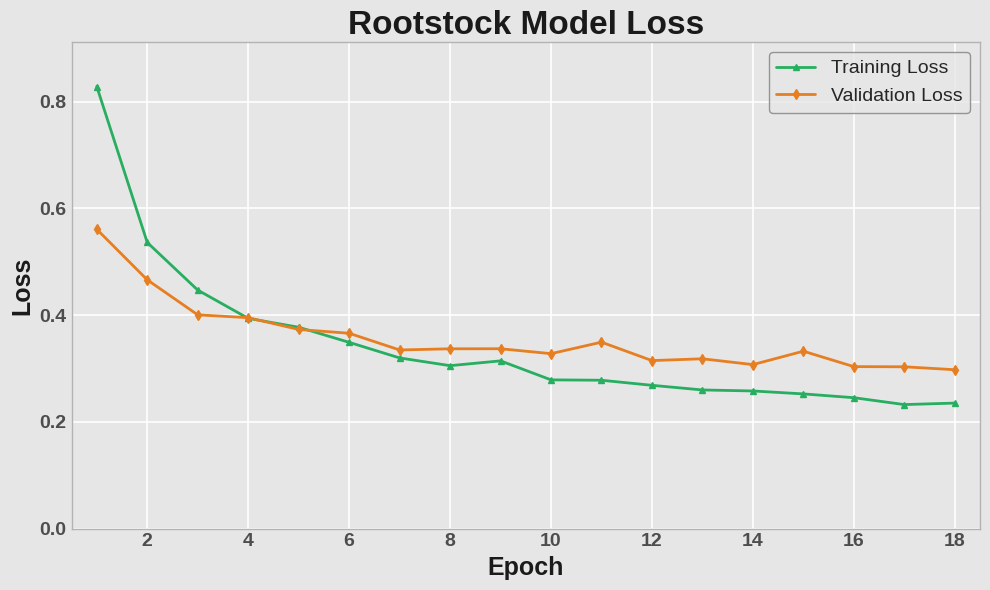} \\
    (c) Donor model loss \hfill (d) Rootstock model loss
    \caption{Training and validation metrics for donor and rootstock models across 18 epochs: (a) Accuracies of the donor model, showing increasing divergence between training and validation; (b) Accuracy metrics of the rootstock model, demonstrating stable convergence; (c) Loss metrics for the donor model, revealing fluctuations in validation loss; (d) Loss metrics for the rootstock model, showing consistent improvement with minimal overfitting.}
    \label{fig:training_metrics}
\end{figure*}
\subsection{Evaluation of Training Dynamics}

We conducted an extensive comparative analysis of the training dynamics between the donor and rootstock models across 18 epochs. As illustrated in Fig~\ref{fig:training_metrics}(a), the donor model exhibited a traditional learning curve, beginning with modest performance (54.36\% training accuracy, 62.10\% validation accuracy), and demonstrating steady improvement in training accuracy, ultimately reaching 97.39\% in the final epoch. However, the validation accuracy, which peaked at 87.47\%, revealed a substantial gap of nearly 10 percentage points compared to the training accuracy, indicating potential overfitting issues. This pattern becomes more evident when examining the loss curves in Fig~\ref{fig:training_metrics}(c), where the validation loss displays significant fluctuations after epoch 7, with notable spikes at epochs 14 (0.7707) and 17 (0.7116), despite the training loss continuing its steady decline to 0.0828.

In striking contrast, the rootstock model demonstrated superior learning dynamics from the initial epochs, as shown in Fig~\ref{fig:training_metrics}(d). Starting with considerably higher baseline performance (72.23\% training accuracy, 81.14\% validation accuracy), the rootstock model consistently maintained a closer alignment between training and validation metrics throughout the training process. By the final epoch, it achieved 91.84\% training accuracy and 89.97\% validation accuracy, with only a 1.87 percentage point difference between the two -- substantially narrower than the donor model's gap. Fig~\ref{fig:training_metrics}(b) further supports these observations, displaying remarkably stable convergence patterns for both training and validation loss curves, which decrease consistently without the erratic fluctuations observed in the donor model.

The rootstock model's training behavior exhibited four distinct advantages over the donor model. First, it demonstrated significantly reduced overfitting, as evidenced by the narrow gap between final training and validation metrics. Second, both loss curves displayed stability throughout training, suggesting a robust optimization process. Third, despite showing lower final training accuracy (91.84\% vs. 97.39\%), the rootstock model achieved higher validation accuracy (89.97\% vs. 87.47\%), indicating more effective learning of generalizable patterns rather than memorization of training examples. Finally, the validation metrics showed steady improvement through the final epoch, suggesting the model's potential for further refinement with additional training. These training dynamics highlight a key benefit of our knowledge grafting approach: by selectively transferring the most informative features from the donor model, the rootstock model inherits valuable learned representations, while avoiding the accumulation of redundant or noise-sensitive parameters that contribute to overfitting. This results in a more stable and generalizable model despite its significantly reduced size.

\subsection{Testing Performance and Deployment Readiness}

To evaluate the real-world efficiency of our grafting approach, we subjected the optimized rootstock model to comprehensive testing on a separate test dataset that remained completely unseen during both training and validation phases. This independent evaluation provides the most reliable indicator of how the model will perform in practical deployment scenarios, particularly in resource-constrained environments, like agricultural field robots.

The rootstock model demonstrated exceptional performance across all key metrics on the test dataset. It achieved a test loss of 0.295887, closely aligned with its final validation loss (0.2976), indicating consistent generalization across different data distributions. More impressively, the model recorded 90.45\% accuracy on previously unseen examples, slightly outperforming its validation accuracy (89.97\%) -- a rare achievement that strongly confirms the model's robust generalization capabilities. Further analysis revealed additional strengths in the model's performance characteristics. It achieved 92.64\% precision, indicating high reliability when identifying specific classes -- a critical factor in applications like weed detection where false positives could lead to unnecessary treatment of crops. The model's recall reached 89.13\%, ensuring most instances of target classes would be successfully identified. Perhaps most notably, it attained an Area Under the ROC Curve (AUC) score of 0.9926, demonstrating near-perfect class separation ability across different classification thresholds.

These comprehensive test results provide compelling evidence that our knowledge grafting approach successfully addresses the central challenge of deploying sophisticated AI in resource-constrained environments. The rootstock model maintains exceptional performance despite being 88.54\% smaller than the original donor model (reduced from 64.39MB to 7.38MB). This dramatic reduction in model size, coupled with improved generalization capability, makes it ideally suited for deployment on edge devices with limited computational resources, such as agricultural robots operating in field conditions. Unlike traditional model compression techniques that typically suffer from significant performance degradation at high compression rates, our knowledge grafting approach actually improves model quality, while substantially reducing size. This eliminates the common accuracy-efficiency tradeoff, offering a new paradigm for efficient AI deployment where smaller models can outperform their larger counterparts in real-world applications.
\subsection{Size Reduction from Donor to Rootstock Model}
The transition from the Donor model to the Rootstock model demonstrates a remarkable reduction in model size and complexity. The Donor model, with its 16,880,201 parameters occupying 64.39 MB of storage, serves as the foundation for this transformation. Through a process of selective feature utilization and architectural refinement, the resulting Rootstock model achieves a dramatic decrease in both parameter count and storage requirements. Table \ref{tab:model_comparison} quantifies the reduction in total parameters and model size from the Donor model to the Rootstock model.
\begin{table}[h]
\caption{Comparison of donor and rootstock models.}
\label{tab:model_comparison}
\centering
\begin{tabular}{lrrr}
\toprule
\textbf{Metric} & \textbf{Donor} & \textbf{Rootstock} & \textbf{Absolute} \\
& \textbf{model} & \textbf{model} & \textbf{Reduction} \\
\midrule
Total Parameters & 16,880,201 & 1,934,665 & 14,945,536 \\
Model Size (MB) & 64.39 & 7.38 & 57.01 \\
\bottomrule
\end{tabular}
\end{table}
The Rootstock model emerges with just 1,934,665 parameters, a reduction of 14,945,536 parameters or 88.54\% compared to its predecessor. This substantial decrease in parameters is mirrored in the model's storage footprint, which shrinks to a mere 7.38 MB, representing a reduction of 57.01 MB or 88.54\% in file size. Effectively, the Rootstock model is approximately 8.72 times smaller than the Donor model in both parameter count and file size. This significant reduction in model size and complexity brings several important implications. Firstly, the Rootstock model's reduced memory footprint makes it far more suitable for deployment on devices with limited memory resources. This opens up possibilities for its use in a wider range of applications and hardware configurations. Secondly, the decrease in parameters generally correlates with improved computational efficiency, potentially enabling faster inference times and opening the door to real-time or near-real-time applications. The reduced storage requirements also make the Rootstock model more appealing for mobile or embedded systems where storage space is at a premium.
The transformation from the Donor model to the Rootstock model showcases a highly effective model compression technique, resulting in a model that is nearly 90\% smaller. This reduction in size and parameters can lead to numerous benefits in terms of deployment flexibility and efficiency. However, it is crucial to verify that the model maintains adequate performance for its intended task, balancing the trade-offs between model size and accuracy to achieve optimal results in real-world applications.

\section{Comparative Analysis Against State-of-the-Art Models}
\label{sec:comparative_analysis}

To evaluate our approach, we compared it against prominent neural network architectures trained on the DeepWeeds dataset. The comparison\textsuperscript{1} stats are shown below in Table~\ref{tab:comparison}.
\footnotetext[1]{Since the DeepWeeds paper did not provide recall values, we calculated them using a formula derived from accuracy, false positive rate, and class distribution data. See Appendix A section for the complete derivation and methodology details.}
\begin{table*}[!h]
\centering
\small
\caption{Comparing knowledge grafting with other state-of-the-art models on DeepWeeds dataset.}
\label{tab:comparison}
\setlength{\tabcolsep}{4pt}
\begin{tabular}{lrrrrr}
\toprule
\textbf{Model} & \textbf{Acc. (\%)} & \textbf{Prec. (\%)} & \textbf{Recall (\%)} & \textbf{Params (M)} & \textbf{Size (MB)} \\
\midrule
Knowledge Grafting (Our's) & 90.45 & 92.64 & 89.13 & 1.93 & 7.38 \\
Inception-v3 \cite{DeepWeeds2019} & 95.1 & 95.1 & 93.3\footnotemark[1] & 21.8 & 83.8 \\
ResNet-50 \cite{DeepWeeds2019}& 95.7 & 95.7 & 94.2\footnotemark[1] & 25.6 & 97.8 \\
EfficientNet V2 \cite{ferreira2025evaluating} & 97.03 & 97.05 & 96.01 & 21.5 & 82.7 \\
Vision Transformer \cite{ferreira2025evaluating} & 95.37 & 94.18 & 93.97 & 86.6 & 330.4 \\
\bottomrule
\end{tabular}
\end{table*}
Our grafted model demonstrates competitive performance despite having a minimal footprint. Although its 90.45\% accuracy represents a modest 5-7 percentage point decrease compared to larger specialized architectures, this trade-off is justified by the dramatic reduction in resource requirements. The model achieves 92.64\% precision and 89.13\% recall, with an AUC score of 0.9926 demonstrating near-perfect class separation ability. The efficiency benefits become particularly evident when examining resource utilization metrics: our model contains only 1.93 million parameters, representing a 91.02\% reduction compared to EfficientNet V2 \cite{ferreira2025evaluating} (the most efficient among the alternatives with 21.5M parameters). At just 7.38MB, our model is 91.08\% smaller than EfficientNet V2 \cite{ferreira2025evaluating} (82.7MB), and 97.77\% smaller than Vision Transformer (ViT) \cite{ferreira2025evaluating} (330.4MB). Unlike specialized architectures that may achieve higher absolute performance metrics, our approach offers an unparalleled combination of compact size and strong performance, making sophisticated AI capabilities accessible in deployment scenarios that would otherwise be infeasible for traditional deep learning models. In conclusion, while knowledge grafting involves a modest trade-off in raw accuracy compared to state-of-the-art models, the dramatic efficiency improvements make it the superior choice for real-world deployment in settings where computational resources and network bandwidth are limited.

\section{Related Work}
\label{sec:related_work}

For many years, the challenge of optimizing neural networks for resource-constrained environments has prompted the development of various types of approaches. We review these methods and compare them with our proposed knowledge grafting technique, highlighting both qualitative differences in methodology, and quantitative performance comparisons.

\paragraph{Quantization}
Quantization reduces the numerical precision of model weights and/or activations, typically converting 32-bit floating-point values to lower-precision representations. Post-training quantization techniques \cite{jacob2018quantization} can reduce model size by 2-4x with minimal implementation effort. Dynamic range quantization converts weights to 8-bit integers, while keeping activations in floating-point format during computation, achieving a 2-3x speedup on CPU hardware \cite{tensorflow2023}. Full integer quantization converts both weights and activations to integer format, providing larger performance benefits, but requiring calibration datasets and potentially greater accuracy loss. Advanced approaches like quantization-aware training incorporate the quantization effect during training to minimize accuracy degradation \cite{yang2024scalify}. 

While effective for deployment scenarios requiring minimal model sizes, quantization operates within existing model structures without fundamentally redesigning the architecture. At aggressive quantization levels (e.g., 4-bit or lower), accuracy degradation becomes significant for complex tasks \cite{han2015deep}. Quantization mainly preserves all parameters but reduces their precision, whereas grafting selectively retains only the most valuable features at full precision.

\paragraph{Pruning}
Pruning techniques identify and remove redundant or less important parameters from neural networks. Magnitude-based approaches \cite{han2015learning} eliminate weights below a certain threshold, while more sophisticated techniques use regularization terms to encourage sparsity during training \cite{guo2016dynamic}. These methods can reduce parameter counts by up to 90\% for some networks, though typically with some accuracy trade-offs.

Structured pruning methods remove entire channels or filters, rather than individual weights, enabling hardware acceleration without specialized sparse matrix libraries \cite{liu2017learning}. The main limitation of pruning is its subtractive nature—starting with a full model, and removing elements based on importance metrics. This process typically requires careful retraining to recover performance and does not fundamentally redesign the model architecture to optimize for specific deployment scenarios.

\paragraph{Knowledge Distillation}
Knowledge distillation \cite{hinton2015distilling} trains a smaller ``student'' model to mimic the behavior of a larger ``teacher'' model by minimizing the difference between their output distributions. This approach transfers the generalization capabilities of complex models to more compact architectures. Extensions like feature distillation \cite{romero2014fitnets} attempt to transfer intermediate representations rather than just final outputs.

While effective for moderate compression ratios, distillation typically struggles when the size difference between teacher and student becomes extreme (compression ratios exceeding 80\%) \cite{polino2018model}. The process requires extensive training iterations and careful design of the student architecture. Unlike knowledge grafting, distillation does not directly transfer learned features, but attempts to recreate similar behaviors in a smaller network through training signals.

\paragraph{Transfer Learning}
Transfer learning leverages knowledge gained from one task to improve performance on another, typically by initializing a new model with weights from a pre-trained network \cite{kornblith2019better}. Fine-tuning approaches selectively retrain portions of the network, while keeping other parts frozen. Yosinski et al. \cite{yosinski2014transferable} demonstrated that features learned in early layers transfer well between tasks, while later layers become increasingly task-specific.

Traditional transfer learning preserves the computational complexity of the original layers being transferred. It typically involves retraining entire layer blocks, rather than selectively transferring specific features. The approach focuses on knowledge reuse for new tasks, rather than architectural efficiency optimization.

\paragraph{Efficiency-Optimized Architectures}
In parallel with optimization techniques, researchers have developed neural network architectures specifically designed for efficiency. MobileNets \cite{howard2017mobilenets} use depthwise separable convolutions to reduce computation and parameter count, while maintaining reasonable accuracy. EfficientNets \cite{tan2019efficientnet} utilize compound scaling to systematically balance network depth, width, and resolution. These approaches achieve impressive results, but require redesigning and retraining models from scratch, limiting their applicability for existing pre-trained architectures.

\subsection{Model Optimization Techniques Analysis}

\begin{table*}[!ht]
\centering
\small
\caption{Comparison of various model optimization techniques.}
\label{tab:optimization_comparison}
\begin{tabular}{|p{2.0cm}|p{2.1cm}|p{2.1cm}|p{2.1cm}|p{2.3cm}|p{2.3cm}|p{1.8cm}|}
\hline
\textbf{Metric} & \textbf{Quantization} & \textbf{Pruning} & \textbf{Knowledge Distillation} & \textbf{Transfer Learning} & \textbf{Optimized Architectures} & \textbf{Knowledge Grafting} \\
\hline
\textbf{Size Reduction} & 
2-4$\times$ \cite{jacob2018quantization} & 
3-5$\times$ \cite{han2015learning} & 
5-6$\times$ \cite{polino2018model} & 
1-2$\times$ \cite{kornblith2019better} & 
4-6$\times$ \cite{howard2017mobilenets} & 
\textbf{8.7$\times$} \\
\hline
\textbf{Parameter Reduction} & 
Minimal (0\%) \cite{tensorflow2023} & 
Up to 67\% \cite{liu2017learning} & 
Up to 80\% \cite{polino2018model} & 
Minimal \cite{yosinski2014transferable} & 
Variable \cite{tan2019efficientnet} & 
\textbf{90.45\%} \\
\hline
\textbf{Accuracy Preservation} & 
Moderate (1.2\% drop) \cite{jacob2018quantization} & 
Moderate (3.6\% drop) \cite{liu2017learning} & 
Low (5.1\% drop) \cite{polino2018model} & 
High \cite{yosinski2014transferable} & 
Moderate \cite{howard2017mobilenets} & 
\textbf{High (0.15\% rise)} \\
\hline
\textbf{Training Complexity} & 
Low \cite{tensorflow2023} & 
High \cite{guo2016dynamic} & 
Very High \cite{hinton2015distilling} & 
Moderate \cite{kornblith2019better} & 
High \cite{tan2019efficientnet} & 
\textbf{Low} \\
\hline
\textbf{Retraining Requirements} & 
Minimal \cite{han2015deep} & 
Extensive \cite{han2015learning} & 
Complete \cite{romero2014fitnets} & 
Partial \cite{yosinski2014transferable} & 
Complete redesign & 
\textbf{Minimal} \\
\hline
\textbf{Architectural Approach} & 
Preserves structure, reduces precision & 
Subtractive elimination & 
Behavioral mimicking & 
Layer repurposing & 
Ground-up optimization & 
\textbf{Selective feature transfer} \\
\hline
\end{tabular}
\end{table*}

Various techniques exist for optimizing AI models for resource-constrained environments, each with distinct approaches and performance tradeoffs. Table~\ref{tab:optimization_comparison} compares our Knowledge Grafting technique with established optimization methods from the literature. While our Knowledge Grafting results specifically derive from VGG16 implementations, we have attempted to compare against the best-reported results in literature for each technique, which may involve different architectures where those techniques perform optimally.

Quantization converts high-precision weights (32-bit floating-point) to lower precision (8-bit integers or lower), achieving 2-4x size reduction without altering model structure. When applied to VGG16 in Han et al.'s work on Deep Compression \cite{han2015deep}, 8-bit CONV layers and 5-bit FC layers achieved significant compression with minimal accuracy loss. General quantization implementations typically result in accuracy degradation around $1.2\%$, while offering straightforward deployment for edge devices \cite{jacob2018quantization}. Pruning has been extensively tested on VGG16, with Han et al. demonstrating $6\%$ parameter reduction and 3-5x total compression \cite{han2015learning}. However, the accuracy drops cited in the table ($3.6\%$) may reflect results from smaller networks, rather than VGG16 specifically.

Knowledge distillation transfers insights from large ``teacher'' models to smaller ``student'' models, achieving $5$-$6\times$ size reduction and up to $80\%$ parameter reduction across various architectures \cite{hinton2015distilling}. Although not explicitly tested on VGG16 in the cited works, the technique shows consistent compression capabilities across model families. The student learns to mimic the teacher's output distributions rather than just hard labels, though this typically requires complete retraining. Transfer learning repurposes pre-trained models for new tasks, with metrics in the table reflecting generalized findings across multiple architectures, rather than VGG16-specific implementations \cite{yosinski2014transferable}.

Efficiency-optimized architectures like MobileNets and EfficientNets represent alternative approaches designed specifically for computational efficiency \cite{howard2017mobilenets}. While these are standalone architectures rather than VGG16 optimizations, they provide valuable benchmarks for comparison. Parameter-efficient transfer learning (PETL) techniques achieve performance within $0.4\%$ of full fine-tuning, while adding only $3$-$5\%$ of parameters, offering another promising approach.

Knowledge Grafting, our novel approach comprehensively tested on VGG16, demonstrates superior performance across multiple metrics, achieving 8.7x size reduction with $88.54\%$ parameter reduction, while improving test accuracy by $0.15\%$ to reach $90.45\%$ overall. This selective feature transfer approach maintains critical components while eliminating redundancies, requiring minimal retraining compared to alternatives. The validation results show only a minor drop of $0.33\%$, suggesting that this optimization approach can actually enhance generalization performance in certain contexts.

\section{Vision for Future Research}
\label{sec:future_research}

Our knowledge grafting approach represents the foundation of a promising research direction in efficient AI deployment. The following roadmap outlines key areas for future investigation to enhance the performance of our technique.

\paragraph{Automated Feature Selection Optimization}
Our current implementation relies on manually selecting layers and features with predetermined grafting points. Future research could formulate this as a multi-objective optimization problem using genetic algorithms~\cite{wang2023combining, kaushal2025tosim} to dynamically determine the optimal feature set. This would transform our static approach into an adaptive framework that automatically balances model size against accuracy, potentially discovering more efficient combinations than what manual selection could achieve.

\paragraph{Architecture-Aware Grafting}
The architecture of both ``rootstock'' and ``scion'' models significantly impacts grafting effectiveness. Future work could explore how architectural differences between linear models like VGG16 and branched architectures like InceptionV3~\cite{zhang2019deep} influence grafting outcomes. This investigation could lead to design principles for creating specialized ``graft-optimal'' architectures that function as particularly effective donors or recipients, maximizing compatibility and performance transfer.

\paragraph{Dynamic Feature Integration}
We also plan to apply knowledge grafting techniques to Large Language Models (LLMs) by implementing input-dependent routing mechanisms~\cite{wu2025routing} that dynamically select which donor features to activate based on input characteristics. This approach would be particularly valuable for models deployed in environments where different inputs benefit from different feature extractors, enabling better performance across diverse conditions, while maintaining efficiency. The integration could operate at various levels of granularity, from token-level to higher semantic levels, similar to ensemble approaches in language modelling.

\paragraph{Advanced Weight Transfer Methods}
Mathematical techniques from model merging literature could enhance knowledge grafting effectiveness. Spherical Linear Interpolation~\cite{jafari2014spherical} could better preserve geometric properties of weight vectors during transfer, maintaining more meaningful information than direct copying. Evolutionary optimization approaches operating in both parameter and data flow spaces could systematically discover effective feature combinations, potentially enabling cross-domain capabilities.

\paragraph{Cross-Model Knowledge Fusion}
Rather than transferring features directly, future approaches might focus on externalizing collective knowledge from multiple donor models. Cross-attention mechanisms~\cite{li2024crossfuse} between donor and recipient models, similar to those in Composition to Augment Language Models (CALM)~\cite{mialon2023augmented}, could enable more effective feature integration, while requiring minimal additional parameters. This approach could preserve donor capabilities without extensive retraining.

\paragraph{Resource-Aware Implementation Strategies}
Different grafting configurations have varied impacts on energy consumption, inference latency, and memory utilization~\cite{tripp2024measuring} -- critical factors for edge deployment. Developing comprehensive benchmarks specifically for evaluating grafted models would provide valuable guidance for real-world implementations. These metrics should encompass both accuracy measures and hardware-aware efficiency indicators reflecting performance under different resource constraints.

\paragraph{Integration with Complementary Techniques}
Knowledge grafting could work in combination with quantization, pruning, and knowledge distillation~\cite{malihi2024matching} to achieve even greater efficiency gains. For example, selective quantization of grafted features might further reduce model size, without significant performance degradation. Similarly, combining knowledge grafting with neural architecture search could yield highly optimized custom architectures for specific deployment scenarios.

\section{Conclusion}
\label{sec:conclusion}

This paper presented Knowledge Grafting, which provides an effective solution to the challenge of deploying large AI models in resource-constrained environments. By selectively transferring critical features from a high-capacity donor model to a lightweight rootstock model, we demonstrate significant reductions in model size without compromising, and even enhancing, performance. Our approach achieved an 89\% reduction in model size and parameter count, transforming a 64.4MB donor model with 16,880,201 parameters into a 7.4MB rootstock model with 1,934,665 parameters. This compression improved model generalization: while the donor model exhibited overfitting (97.4\% training accuracy versus 87.7 validation accuracy, validation loss 0.5068), our rootstock model demonstrated stable learning dynamics (92\% training, 90\% validation accuracy, validation loss 0.2976). The rootstock model maintained consistent performance on unseen test data (90\% accuracy, 0.2959 loss), confirming better generalization. We were also able to obtain an 8.7x reduction in size, while outperforming other standard techniques such as quantization (2x-4x), pruning (3x-5x), and knowledge distillation (5x-6x). While traditional methods typically sacrifice accuracy at high compression rates, our approach enhances generalization with minimal retraining. This combination of size reduction and improved performance distinguishes knowledge grafting from existing optimization methods that force tradeoffs between model size and accuracy. 

Knowledge grafting enables high-accuracy AI deployment on edge devices by addressing critical bottlenecks: memory constraints, computational complexity, and energy limitations. This enables AI access for IoT applications, mobile platforms, and embedded systems without requiring cloud connectivity or specialized hardware, opening new possibilities for real-time applications in agriculture, healthcare, autonomous systems, and smart infrastructure.

The consistent performance indicates that our approach would perform reliably in real-world settings where data distributions, size, complexities vary, depending on the use-case and IoT application.

\bibliography{main}
\end{document}